\documentclass[conference]{IEEEtran}
\IEEEoverridecommandlockouts
\usepackage{cite}
\usepackage{amsmath,amssymb,amsfonts}
\usepackage{algorithmic}
\usepackage{graphicx}
\usepackage{textcomp}
\usepackage{xcolor}
\def\BibTeX{{\rm B\kern-.05em{\sc i\kern-.025em b}\kern-.08em
    T\kern-.1667em\lower.7ex\hbox{E}\kern-.125emX}}
\begin{document}

\title{LVIC: Multi-modality segmentation by Lifting Visual Info as Cue \\
{\footnotesize \textsuperscript{*}Note: Technical Report Version}
}

\author{\IEEEauthorblockN{1\textsuperscript{st} Zichao Dong}
\IEEEauthorblockA{\textit{UDeer.ai} \\
zichao@udeer.ai}
\and
\IEEEauthorblockN{2\textsuperscript{nd} Bowen Pang}
\IEEEauthorblockA{\textit{Zhejiang University} \\
pangbw@zju.edu.cn
}
\and
\IEEEauthorblockN{3\textsuperscript{rd} Xufeng Huang}
\IEEEauthorblockA{\textit{UDeer.ai} \\
xufeng@udeer.ai
}
\and
\IEEEauthorblockN{4\textsuperscript{th} Hang Ji}
\IEEEauthorblockA{\textit{UDeer.ai} \\
jihang@udeer.ai
}
\and
\IEEEauthorblockN{5\textsuperscript{th} Xin Zhan}
\IEEEauthorblockA{\textit{UDeer.ai} \\
zhanxin@udeer.ai}
\and
\IEEEauthorblockN{6\textsuperscript{th} Junbo Chen *}
\IEEEauthorblockA{\textit{UDeer.ai} \\
junbo@udeer.ai}

\thanks{This is a draft version for now.}

}

\maketitle

\begin{abstract}  

Multi-modality fusion is proven an effective method for 3d perception for autonomous driving. However, most current multi-modality fusion pipelines for LiDAR semantic segmentation have complicated fusion mechanisms. Point painting is a quite straight forward method which directly bind LiDAR points with visual information. Unfortunately, previous point painting like methods suffer from projection error between camera and LiDAR. In our experiments, we find that this projection error is the devil in point painting. As a result of that, we propose a depth aware point painting mechanism, which significantly boosts the multi-modality fusion. Apart from that, we take a deeper look at the desired visual feature for LiDAR to operate semantic segmentation. By Lifting Visual Information as Cue, LVIC ranks 1st on nuScenes LiDAR semantic segmentation benchmark. Our experiments show the robustness and effectiveness. Codes would be make publicly available soon. 

\end{abstract}

\begin{IEEEkeywords}
Point cloud recognition, LiDAR semantic segmentation, multi-modality semantic segmentation.
\end{IEEEkeywords}

\section{Introduction}
Semantic segmentation is a most common and effect way for scene parsing in autonomous driving scenes. For most autonomous driving vehicles, multiple sensors are equipped in order to provide adequate raw data for perception. As we all know, LiDAR does good in providing precise geometric data while camera is suitable to provide detail and textual information. Thus, it is natural to compensate LiDAR with calibrated camera as a better new sensor. However, even for well calibrated LiDAR camera pairs, mis-projected pixel and LiDAR point pairs still common seen. We assume that below issues would inevitably cause projection error. Firstly, calibration would never be perfect itself especially for points far from sensor. Secondly, diverse sensor are installed in different location which would bring dissimilar view. It is hard for naive point painting models to distinguish whether the painted visual information is valid. We notice that difference between depth estimated by vision and LiDAR points is a simple yet effective way to provide the confidence score. Nevertheless, we notice that previous attempts on point painting mainly use object-level result or high level feature embedding. In our experiments, we found that low-level appearance feature is better for LiDAR painting. 

Our contributions can be summarized as follows:  

1. A more simple and effective point painting method is designed for multi-modality semantic segmentation.

2. A depth aware module is added for LiDAR points to decide whether to use information provided by visual patches.

3. Without bells and whistles, our LVIC achieves state-of-the-art performance on nuScenes LiDAR semantic segmentation leaderboard.

\section{RELATED WORK}
\subsection{Point cloud segmentation}
Point cloud segmentation is the process of predicting the category of every point in a set of point clouds. In outdoor scenes, point clouds generated by LiDAR present a specific segmentation challenge known as LiDAR-based 3D segmentation. This task entails classifying point clouds based on their semantic attributes. Existing approaches can be classified into three main streams depending on how point clouds are represented: point-based methods \cite{qi2017pointnet,qi2017pointnet++}, voxel/grid-based methods \cite{cciccek20163d,milioto2019rangenet++}, and multi-modal methods \cite{xu2021rpvnet,tang2020searching}. Point-based methods operate on individual points, while voxel/grid-based methods discretize point clouds into 3D voxels or 2D grids and apply convolutional operations. Our approach, LVIC, belongs to the multi-modal method category. In our study, we propose a novel multi-modality fusion method by lifting visual information as cue.

\subsection{Point Transformers}
The Point Transformer series of papers constitutes a significant contribution to the field of point cloud processing. Zhao \textit{et al.} \cite{zhao2021point} contructed Point Transformer network based on the Point Transformer layer, which is invariant to permutation and cardinality. Adopting the PointNet++ \cite{qi2017pointnet++} hierarchical architecture, it replaced the MLP modules with local Transformer blocks. Each block was applied on K-Nearest Neighbor (KNN) neighborhoods of sample points. Vector attention \cite{zhao2020exploring} was used instead of the scalar attention, which has been proven to be more effective for point cloud processing. Point Transformer V2 (PTv2) \cite{wu2022point} improves Point Transformer \cite{zhao2021point} by proposing Grouped Vector Attention (GVA) while designing a partition-based pooling strategy to make network more efficient and spatial information better-aligned.

Point Transformer V3 (PTv3) \cite{wu2024ptv3} aimed to overcome the trade-offs between accuracy and efficiency in point cloud processing. It employed point cloud serialization to transform unstructured point clouds into a structured format, decreasing the time KNN query occupied. For serialized point clouds, Wu \textit{et al.} \cite{wu2024ptv3} designed a streamlined approach. Leveraging the potential of scalability, compared to PTv2, PTv3 achieved higher inference speed, less memory usage while better results across over 20 tasks.

\subsection{Depth estimation}
Depth estimation is key to inferring scene geometry from 2D images. Traditional depth estimation methods are usually based on stereo cameras which rely on point correspondences between images and triangulation to estimate the depth, typically based on hand-crafted \cite{flynn2016deepstereo} or learned features \cite{smolyanskiy2018importance}. However, the stereo depth estimation method requires at least two fixed cameras, and it is difficult to capture enough features in the image to match when the scene has less texture \cite{ming2021deep}. 

Given a single RGB image as input, monocular depth estimation aims to predict the depth value of each pixel, which does not require additional complicated equipments. Supervised single-image depth estimation methods can be categorized into regressing metric depth and relative depth. For absolute depth estimation, Li \textit{et al.} \cite{li2017two} introduced a dual-stream framework based on VGG-16 \cite{simonyan2014very}: one stream for depth regression and the other for depth gradients. These streams were combined using a depth-gradient fusion module to produce a consistent depth map. For relative depth learning, Lee \textit{et al.} \cite{lee2019monocular} formulated a CNN to estimate the relative depth across different scales, which was optimally reorganized to reconstruct the final depth map.

\subsection{ZoeDepth}
As mentioned above, the existing monocular depth estimation approaches either focus on generalization performance while neglecting scale, i.e., relative depth estimation, or concentrate on state-of-the-art results on specific datasets, i.e., absolute depth estimation. ZoeDepth \cite{bhat2023zoedepth} introduced an approach that bridges these two paradigms, resulting in a model that achieves outstanding generalization performance while preserving metric scale.

Bhat \textit{et al.} \cite{bhat2023zoedepth} first employed an Encoder-Decoder backbone for relative depth prediction. Inspired by the LocalBins architecture \cite{bhat2022localbins},  they attaches a novel metric bins module to the decoder to predict the bin centers at every pixel. The metric bins module takes multi-scale features from the decoder and implements multi-scale refinement of the bins by attracting strategy instead of splitting operation in LocalBins. Considering the ordered bins, they used a binomial distribution instead of softmax to predict the probability distribution over the bin centers. Training an absolute depth model on a mixed dataset with various scenes can be challenging. Bhat \textit{et al.} addresses this by pretraining a backbone for relative depth estimation, which alleviates the fine-tuning issue across multiple datasets to some extent. Subsequently, the model is equipped with multiple Metric bins modules, each corresponding to a scene type (indoor and outdoor). Finally, end-to-end fine-tuning is conducted on the complete model.

\subsection{Multi-modality semantic segmentation}
Most multi-modal 3D Semantic Segmentation uses LiDAR and camera to improve perception ability. The camera provides RGB and dense context information, while LiDAR provides more spatial and geometric information about the surrounding environment. PMF \cite{zhuang2021perception} projects the point cloud onto the image plane to fuse it with image data, leading to the improved segmentation performance within the field of view (FOV). MSeg3D \cite{li2023mseg3d} proposed a multi-modal 3D semantic segmentation model that combined intra-modal feature extraction and inter-modal feature fusion, which to some extent mitigated the modal heterogeneity. 2D3DNet \cite{genova2021learning} proposed an approach to train a 3D model from pseudo-labels derived from 2D semantic image segmentations using multiview fusion.

\subsection{Previous point painting method}
PointPainting \cite{vora2020pointpainting} operates by projecting LiDAR points onto the output of an image-only semantic segmentation network and then attaching class scores to each point. This augmented (painted) point cloud can subsequently be utilized as input for any LiDAR-only method.
MVP \cite{yin2021multimodal} designed a straightforward framework for fusing 3D LiDAR data with high-resolution color images. Yin \textit{et al.} \cite{yin2021multimodal} utilizes close-range depth measurements from the LiDAR sensor to project RGB pixels into the scene, thereby  converting RGB pixels into three-dimensional virtual points. This multi-modal approach enables MVP to produce high-resolution three-dimensional point clouds near target objects. Subsequently, a center-based 3D detector is employed to identify all objects within the scene.

\subsection{CPGNet-LCF}
The fusion of LiDAR and camera always needs an accurate calibration matrix from LiDAR to cameras. However, most of the existing multi-modal methods did not take this into consideration. Meanwhile, the fusion models usually meet the difficulty with efficient deployment and real-time execution.To address these challenges, Jiang \textit{et al.} \cite{jiang2023revisiting} proposed CPGNet-LCF based on CPGNet \cite{li2022cpgnet}, which can be easily employed and capable of performing real-time inference. CPGNet-LCF \cite{jiang2023revisiting} used STDC \cite{fan2021rethinking} as the image backbone to extract rich texture information. LiDAR point features are augmented by bilinear-sampling the image features and then undergo the LiDAR segmentation backbone, which is a BEV and RV fusion framework following CPGNet, to produce the final segmentation results.
During training, CPGNet-LCF used a novel weak calibration knowledge distillation strategy to improve the robustness against the weak calibration. Specifically, Jiang \textit{et al.} adds the noise in the weak calibration matrix as a kind of data augmentation. The model trained by the well calibration samples serves as the teacher model to guide the student model trained by weak calibration samples, so that the student model is robust against the weak calibration.

\section{METHOD}
\subsection{Overview}
\begin{figure*}[htbp]
    \centering
    \includegraphics[width=16cm]{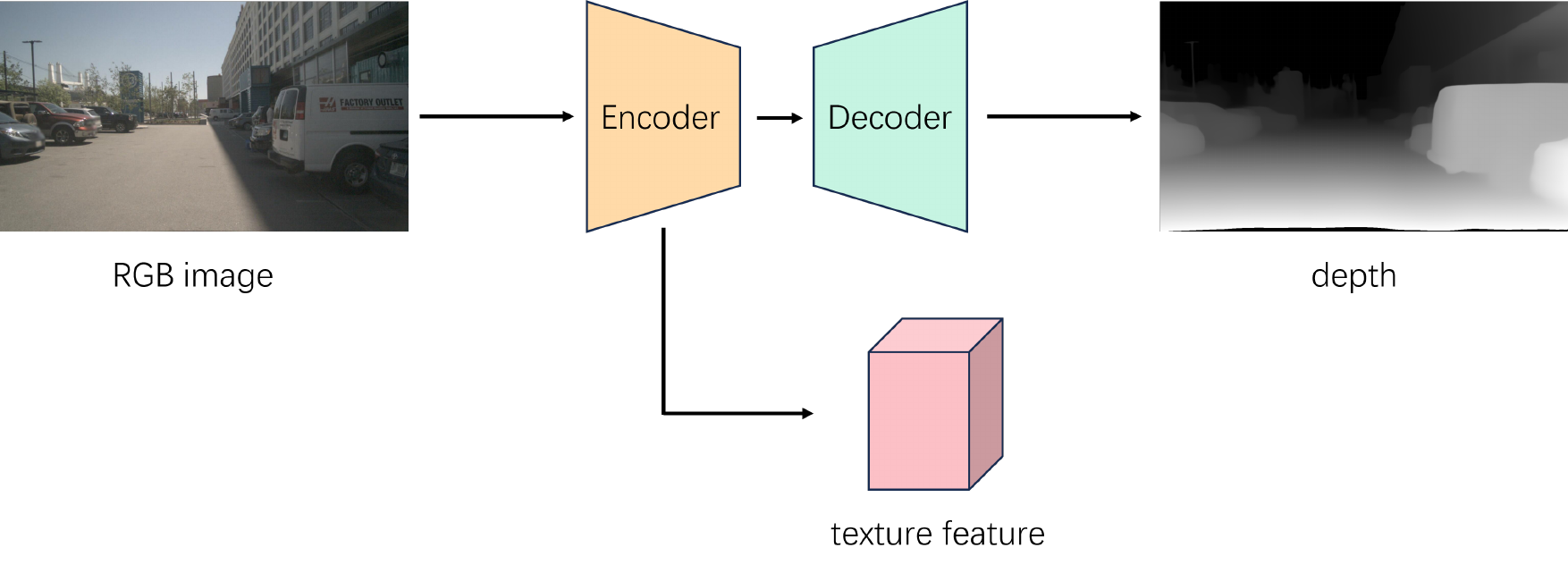}
    \caption{\small \textbf{Pipeline of our visual encoder}. The whole model is arbitrarily a encoder decoder architecture. Low level feature would be saved during encoder.}
    \label{fig:model}
\end{figure*}

Our LVIC is mainly constructed by three main components, visual encoder, painting module and fusion module. Below section would depict them in detail. The pipeline of LVIC is shown in Fig.~\ref{fig:model}. Details and experiments are illustrated in below sections. 

\subsection{Visual encoder}
Speaking of point painting, extracting valuable feature from visual input is the most important issue. We argue that there are two main aspects lying here. Firstly, since LiDAR does pool in texture feature, dense and detail features are needed. Secondly, due to inevitable projection error, a token representing confidence score of projection accuracy is the key of camera LiDAR fusion. We believe that the latter point is the key to make point painting great again.  

Firstly, we investigate in optimal feature for painting. As for the architecture of visual encoder, there are two directions. On the one hand, nowadays ViTs\cite{dosovitskiy2020image} shows strong ability in multiple visual tasks due to his data dependant characteristic and global long range reception field. On the other hand, CNN based models like ResNet\cite{DBLP:journals/corr/HeZRS15} and ConvNexts\cite{liu2022convnet}\cite{woo2023convnext} good at dense modeling especially for segmentation tasks, while adapters like ViT-adapters should be added for ViTs to acquire dense features. Our experiments shows that low-level feature is good enough for LiDAR painting. However, for high efficiency in painting step where painting result would be saved locally, we choose EfficientViT\cite{cai2022efficientvit} as our visual encoder. 

Secondly, projection error is also expected be solved by visual encoder. Assuming in camera coordination, a LiDAR point has position (x\_l, y\_l, z\_l) who is projected to a pixel with 2d coordination (u, v) according to the calibration parameters. However, even though they are considered as matched pairs by calibration relationship, they may represent diverse regions. To be specific, the pixel located in (u, v) should corresponding to (x\_c, y\_c, z\_c) with depth z\_c. Thanks to recent excellent works focusing depth estimation, above z\_c could be estimated properly. Thus, for 3d point (x\_l, y\_l, z\_l) we could provide the texture feature associated with his calibrated pair with (x\_c, y\_c, z\_c) indicating the 3d affinity. 

In summary, the vision encoder is typically a EfficientViT with depth estimation decoder. Input a monocular image with shape H*W, a feature map with shape H/4 * W/4 representing texture feature by 4*4 patches and a depth map with shape H*W are generated. Detail settings of visual encoder could be seen in following section.  

\subsection{Painting module}
In our pipeline, painting module is offline and isolated with visual encoder module and point cloud segmentation module. For each point in point cloud set, we first project it to corresponding pixel according to calibration results. Notably, for points could not be projected to any valid pixel, -1 is padded for all painting dimensions as a marker. Apart from that, a LiDAR point with coordination (x, y, z) would be painted with corresponding pixel 2d coordination (u, v), depth estimation result z\_c and a d-dimension texture feature.  

Each painted dimension are concatenated with original points. For the original point cloud set with shape [n, c] where n stands for number of points and c stands for the original feature dimension of each point. After point painting module, the point cloud would have shape [n, c + 3 + 1 + d]. 

Intuitively, a point located in (x, y, z) only have weak features like intensity at the beginning.After painting, appearance feature is added which would carry out more useful information. 

\subsection{Fusion module}
\begin{figure*}[htbp]
    \centering
    \includegraphics[width=14cm]{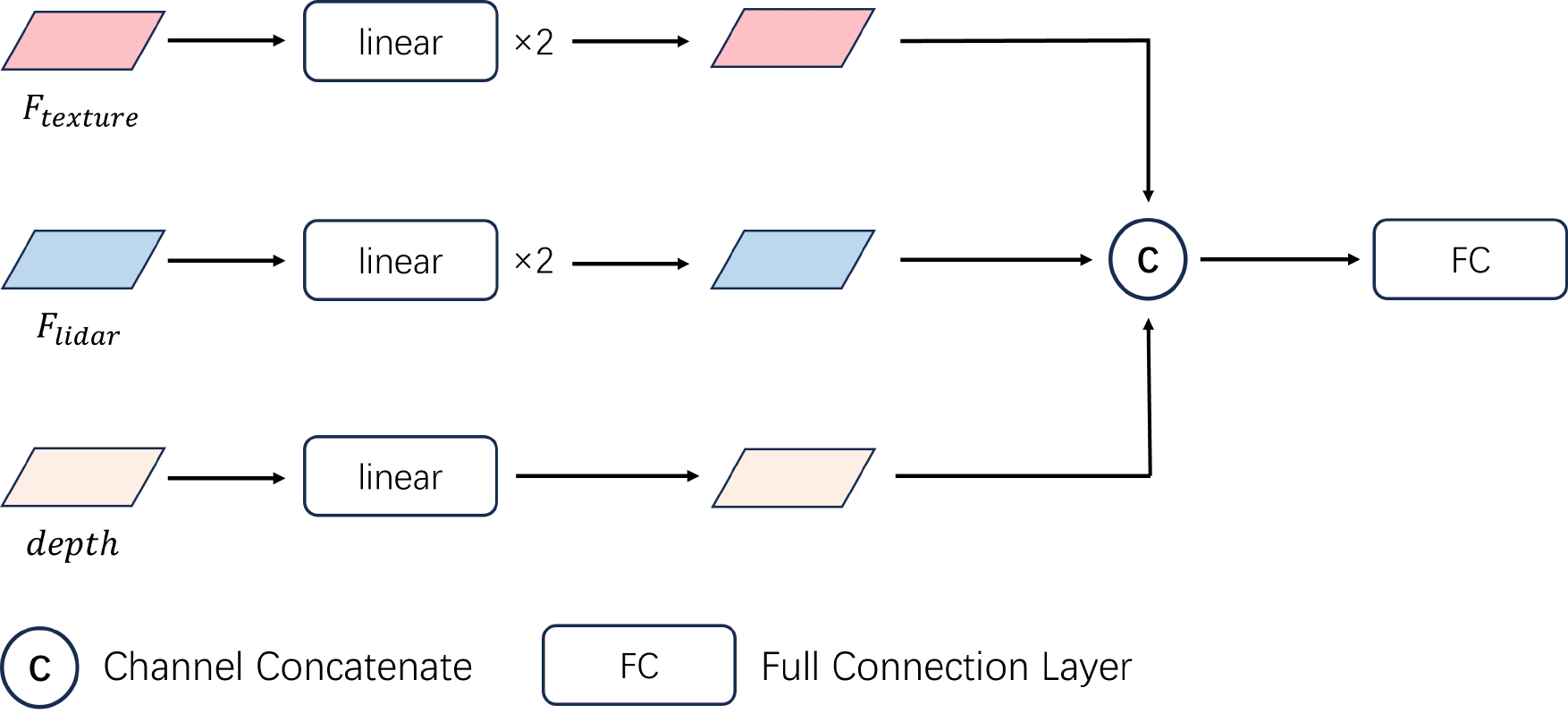}
    \caption{\small \textbf{Components of fusion model}. We only use simple linear layer as adaptor to fuse feature from multiple domain. }
    \label{fig:fusion}
\end{figure*}

Fusion model is a plug-and-play module which is friendly to most existence point cloud related deep learning models. For most point cloud models, the first step is point embedding, which would embedding point-level feature. To be concrete, taking LiDAR frame as input, a embedding module would project it to a feature vector. As illustrated in Fig.~\ref{fig:fusion}, our fusion model is integrated as a point embedding module.

Similar to LLava\cite{liu2023llava}, two linear layers are first deployed to adapt visual texture feature for LiDAR. In parallel, a projection layer is used to project (x, y, z) to LiDAR geometry feature. Finally, a fully connection layer is used to fuse geometry feature with texture feature according to painted depth cue. Detail parameter would be seen in following section.

Compared with cross-attention based fusion method like BEVFormer\cite{li2022bevformer} for object detection and CPG-LCF\cite{li2022cpgnet} for semantic segmentation, our method is quite simple while effective. 

\subsection{LiDAR semantic segmentation model}
We use UDeerPep\cite{dong2024pep} as our base LiDAR segmentation model. Various base models are also tried with on-par or equal improvement. Detailed metric could be seen in the following section. 

\section{Experiment}

\subsection{Dataset}
We validate the proposed LVIC on nuScenes LiDAR semantic segmentation dataset.

\subsubsection{nuScenes LiDAR semantic segmentation}
nuScenes \cite{caesar2020nuscenes} stands as a well-known multimodal dataset tailored for 3D object detection within urban environments. Comprising 1000 distinct driving sequences, each spanning 20 seconds, the dataset boasts meticulous 3D bounding box annotations. The LiDAR operates at a frequency of 20Hz, and while it offers sensor and vehicle pose data for each LiDAR frame, object annotations occur only every ten frames, equating to a 0.5-second interval. The dataset goes to great lengths to safeguard privacy, obscuring any personally identifiable information and pixelating faces and license plates in the color images. Impressively, it incorporates a total of six RGB cameras, boasting a resolution of 1600 × 900 and capturing data at a rate of 12Hz. As for LiDAR semgemntation task, we follow panoptic nuScenes \cite{fong2021panoptic} setting.

\subsection{Implementation Details}
\subsubsection{Visual encoder} For visual backbone, EfficientVit-b0 is used. With image input size 3 * H * W, texture feature with shape 16 * H//4 * W//4 is saved. Depth map result with shape 1 * H *W is also acquired as well. We use SGD with learning rate 1e-3.  

\subsubsection{Fusion module} For visual adapter, the embedding dimension are 8 and 4 for two linear layers. We use GELU\cite{DBLP:journals/corr/HendrycksG16} as our activation function. For LiDAR point encoder, the embedding dimension are 4 and 4 for two linear layers. We also use GELU as our activation function.

\subsection{Quantitive Evaluation}
The quantitive results on NuScenes dataset are shown in Table \ref{table:table2}. 

\begin{table*}[htbp]
\centering
\caption{The results on NuScenes test set with TTA.}
\label{table:table2}
\vspace{4mm}
\centering
\tabcolsep 3pt
\begin{tabular}{lccccccccccccccccc}
\hline
\textbf{}                       & \textbf{mIOU} & \rotatebox{90}{\textbf{barrier}} & \rotatebox{90}{\textbf{bicycle}} & \rotatebox{90}{\textbf{bus}} & \rotatebox{90}{\textbf{car}} & \rotatebox{90}{\textbf{construction}} & \rotatebox{90}{\textbf{motocycle}} & \rotatebox{90}{\textbf{pedstrain}} & \rotatebox{90}{\textbf{Traffic cone}} & \rotatebox{90}{\textbf{trailer}} & \rotatebox{90}{\textbf{truck}} & \rotatebox{90}{\textbf{drivable}} & \rotatebox{90}{\textbf{Other flat}} & \rotatebox{90}{\textbf{sidewalk}} & \rotatebox{90}{\textbf{terrian}} & \rotatebox{90}{\textbf{manmade}} & \rotatebox{90}{\textbf{vegetation}} \\ \hline
\textbf{Baseline(Pep)} & 81.8          & 85.5             & 55.5             & 90.5         & 91.6         & 72.7                  & 85.6               & 81.4               & 76.3                  & 87.3             & 74.0           & 97.7                & 70.2                & 81.1              & 77.4               & 92.7               & 90.2                \\
\textbf{Ours}  & 83.8  & 83.7 & 68.7 & 95.8  & 90.3 & 81.5  & 86.6  & 84.3  & 78.0               & 85.7                  & 77.1             & 97.7             & 71.0              & 79.8                & 77.7              & 92.4             & 90.2                               \\ \hline
\end{tabular}
\end{table*}



\bibliography{references}
\bibliographystyle{plain}


\end{document}